\documentclass[11pt,a4paper,twoside,notitlepage]{report}

\usepackage[T1]{fontenc}
\usepackage[utf8]{inputenc}
\usepackage[vmargin=20mm,hmargin=15mm]{geometry}
\usepackage{graphicx}
\usepackage{booktabs}
\usepackage{amsmath}
\usepackage{amsfonts}
\usepackage{multicol}
\usepackage{balance}
\usepackage{enumitem}
\usepackage{fancyhdr}
\usepackage{titlesec}
\usepackage{setspace}
\usepackage{etoolbox}
\usepackage{caption}
\usepackage[hidelinks]{hyperref}

\newcommand{\spanblock}[1]{%
  \end{multicols}%
  \par\bigskip
  {\centering #1\par}%
  \bigskip
  \begin{multicols}{2}%
}

\setlist[itemize]{itemsep=0pt, topsep=3pt}
\setlist[enumerate]{itemsep=0pt, topsep=3pt}
\makeatletter
\patchcmd{\chapter}
  {\if@openright\cleardoublepage\else\clearpage\fi}{}{}{}
\patchcmd{\chapter}{\thispagestyle{plain}}{}{}{}
\makeatother

\titleformat{\chapter}[block]
  {\normalfont\Large\bfseries}{\thechapter.}{1em}{}
\titlespacing*{\chapter}{0pt}{15pt}{10pt}

\titleformat{\section}[block]
  {\normalfont\large\bfseries}{\thesection}{1em}{}
\titlespacing*{\section}{0pt}{8pt}{4pt}

\titleformat{\subsection}[block]
  {\normalfont\normalsize\bfseries}{\thesubsection}{1em}{}
\titlespacing*{\subsection}{0pt}{6pt}{3pt}

\hypersetup{
  pdftitle={T-MoXAI: A Hierarchical Explainability Framework
            for Temporal Multimodal Data},
  pdfauthor={Ali Inha, Mo Vali, Saaliha Vali, Pietro Liò, Meen-Yau Thum}
}

\begin{document}
\begin{center}
  {\LARGE\bfseries
    T-MoXAI: A Hierarchical Explainability Framework\\[0.2em]
    for Temporal Multimodal Data\par}
  \vspace{0.9em}
  {\large
    Ali Inha\textsuperscript{1},
    Mo Vali\textsuperscript{2},
    Saaliha Vali\textsuperscript{3},
    Pietro Li\`o\textsuperscript{1},
    Meen-Yau Thum\textsuperscript{4}\par}
  \vspace{0.65em}
  {\small
    \textsuperscript{1}Department of Computer Science, University of Cambridge, Cambridge, UK\par
    \textsuperscript{2}Cavendish Laboratory, Department of Physics, University of Cambridge, Cambridge, UK\par
    \textsuperscript{3}Imperial College Healthcare NHS Trust, London, UK\par
    \textsuperscript{4}Lister Fertility Clinic, The Lister Hospital, HCA UK, UK\par}

\end{center}

\vspace{0.5em}

\begin{abstract}
\noindent
Artificial Intelligence (AI) models that process temporal multimodal data show considerable potential in high stakes domains such as healthcare and agriculture; however, their inherent opacity can limit trust and practical adoption. We introduce T-MoXAI (Temporal Multimodal eXplainable AI), a hierarchical framework designed to provide explanations at three distinct levels: (1) \emph{when} specific timepoints most strongly influence predictions, identified through temporal Shapley values; (2) \emph{which} data modalities contribute most significantly at those critical moments, determined via attention analysis; and (3) \emph{what} specific features or image regions drive the model's decisions, identified using gradient based attribution methods. Built on a transformer based architecture that effectively handles irregular temporal sequences and heterogeneous data types, T-MoXAI generates all three levels of explanation in under one second, making it suitable for interactive decision support. We evaluate the framework's cross domain applicability on two challenging real world datasets: predicting IVF treatment outcomes from ultrasound sequences and clinical measurements (achieving an AUC of 0.660 despite significant class imbalance), and forecasting wheat yield from temporal RGB imagery and phenotypic traits (achieving an $R^2$ of 0.265 in the presence of substantial environmental variability). Ablation studies indicate that temporal modelling is a critical component for both domains, as its removal leads to performance equivalent to random guessing. Our faithfulness evaluation using temporal ROAR experiments provides evidence that the generated explanations accurately reflect the model's reasoning process. By offering a unified, domain agnostic architectural pattern with an open source implementation, T-MoXAI provides a solid baseline for temporal multimodal XAI, seeking to address the current fragmentation in the field and facilitate the deployment of trustworthy AI in applications where understanding model decisions is as important as predictive accuracy.
\end{abstract}

\chapter{Introduction}

\begin{multicols}{2}

The rapid advancement of deep learning has significantly enhanced the ability to analyse complex, high dimensional data across numerous domains. In healthcare, AI systems have shown notable capabilities in tasks such as segmenting tumours in MRI scans, detecting diabetic retinopathy from fundus photographs, predicting patient deterioration from electronic health records, and recommending personalised treatment strategies \cite{tjoa2021}. Similarly, in precision agriculture, AI models are increasingly used for tasks like forecasting crop yields from satellite imagery, identifying plant diseases from drone photographs, optimising irrigation schedules based on sensor data, and predicting optimal harvest times from weather patterns \cite{kamilaris2018}. These achievements stem from the ability of deep learning to automatically discover intricate, nonlinear patterns within large datasets, often identifying subtle signals that may elude human detection.

However, this high performance is often accompanied by a significant challenge: a lack of transparency. Modern deep learning models, particularly those employing multiple layers of nonlinear transformations, can operate as "black boxes" whose internal decision making processes remain opaque even to their creators. This opacity creates substantial barriers to adoption in safety critical applications where accountability, trust, and the ability to verify results are paramount \cite{rudin2019, albahri2023}. For instance, a physician may be hesitant to act on an AI's recommendation to alter treatment without understanding the reasoning, as doing so could mask systematic biases or potential failure modes. Similarly, an agronomist may be unable to justify costly interventions like additional fertilisation based solely on an unexplained model prediction. This fundamental tension between performance and interpretability has catalysed the emergence of Explainable AI (XAI), a field dedicated to making the decisions of machine learning models understandable to human users \cite{arrieta2019}.

The challenge of explainability becomes considerably more complex when dealing with temporal multimodal data, a scenario increasingly common in real world applications. Consider the process of in vitro fertilisation (IVF) treatment, where specialists monitor patients through periodic transvaginal ultrasound scans, regular blood tests, and clinical examinations over several weeks. Each data modality provides complementary information: ultrasounds reveal follicular development, blood tests indicate hormonal status, and clinical factors capture patient history. The temporal dimension adds another layer of complexity, as the relevance of these measurements evolves throughout the treatment cycle.

Similarly, in precision agriculture, crop development is tracked using a variety of temporal multimodal data sources. Satellite or drone imagery captures visual indicators of plant health, ground based sensors measure soil conditions, weather stations record environmental data, and field measurements document phenotypic traits. The importance of each data source shifts throughout the growing season: early imagery may indicate emergence uniformity, while late season observations can help predict yield.

Traditional XAI approaches typically face three key limitations when confronted with rich, temporally evolving multimodal data. First, they often focus on single modalities, such as image only methods like Grad-CAM \cite{selvaraju2019} or tabular only techniques like SHAP \cite{lundberg2017}, failing to capture the interplay between different data types. Second, many ignore temporal dynamics, treating each timepoint as independent. Third, current solutions frequently employ domain specific architectures, such as models designed for regular video frames \cite{bertasius2021} or electronic health records \cite{choi2016}, which can prevent generalisation across domains.

This fragmentation in the temporal multimodal XAI landscape presents practical and scientific challenges. It complicates the comparison of different approaches, and researchers may need to solve similar problems in isolation. The lack of unified frameworks hinders the establishment of benchmarks and best practices that could accelerate the field's maturation.

To address these limitations, we introduce T-MoXAI (Temporal Multimodal eXplainable AI), a domain agnostic framework that provides hierarchical explanations for systems processing temporal sequences of heterogeneous data. Our framework is designed to answer three fundamental questions that mirror the reasoning process of domain experts:

\begin{enumerate}
    \item \textbf{When} do specific timepoints in the sequence most strongly influence the model's prediction?
    \item \textbf{Which} data modality contributes most significantly at those influential timepoints?
    \item \textbf{What} specific features or image regions ultimately drive the decision?
\end{enumerate}

Our key contributions are as follows:

\begin{enumerate}
    \item \textbf{A unified, domain agnostic architecture} that effectively processes temporal sequences combining images and tabular data, configurable through parameter files rather than requiring code modifications for new domains.
    \item \textbf{A hierarchical three level explanation framework} that integrates multiple XAI techniques (temporal Shapley values, attention analysis, and gradient based methods) to provide comprehensive insights.
    \item \textbf{Cross domain validation} demonstrating the framework's applicability on two distinct real world tasks: predicting IVF outcomes and forecasting wheat yield.
    \item \textbf{A reusable open source baseline} that addresses the need for standardised benchmarks in temporal multimodal XAI.
\end{enumerate}

\chapter{Related Work}

\section{Foundations of Explainable AI}
The field of Explainable AI encompasses diverse approaches aimed at making machine learning models more interpretable. These methods can be categorised along several dimensions that clarify their capabilities and limitations \cite{arrieta2019}.

\textbf{Intrinsic vs. Post hoc Explainability:} Intrinsically interpretable models, such as linear regression and decision trees, incorporate transparency into their architecture. In deep learning, attention mechanisms have emerged as a form of intrinsic interpretability \cite{bahdanau2014}. Attention based methods have been frequently adopted in medical AI. The RETAIN model, for instance, used two level attention to assign importance to both historical visits and clinical variables within visits \cite{choi2016}. However, recent research has questioned whether attention weights faithfully represent model reasoning \cite{wiegreffe2019}, which motivates our multi method approach.

Post hoc methods explain already trained models without modifying their architecture. LIME (Local Interpretable Model Agnostic Explanations) approximates complex models locally with interpretable surrogates \cite{ribeiro2016}. SHAP (SHapley Additive exPlanations) offers a principled approach grounded in cooperative game theory, assigning each feature a Shapley value that represents its contribution to a prediction \cite{lundberg2017}.

\textbf{Local vs. Global Explanations:} A further distinction lies in explanation scope. Local explanations focus on understanding individual predictions, while global explanations attempt to characterise overall model behaviour.

\section{Visual Explanation Techniques}
For models processing images, specialised techniques leverage gradients to identify influential regions. Early approaches created saliency maps from raw gradient magnitudes \cite{simonyan2014}. Grad-CAM (Gradient-weighted Class Activation Mapping) has become widely adopted; it weights and combines feature maps from the final convolutional layer to produce coarse heatmaps \cite{selvaraju2019}. Integrated Gradients addresses the issue of gradient saturation by integrating gradients along a path from a baseline to the input, providing more reliable attributions \cite{sundararajan2017}.

\section{Temporal and Multimodal Explanations}
Recent research has begun to address explanations for temporal and multimodal data, though often separately. Temporal XAI focuses on identifying important time periods and patterns \cite{rojat2021}. Multimodal XAI must explain both individual modality contributions and their interactions \cite{joshi2021}. A key challenge is the disparate nature of explanations across modalities (e.g., pixel level vs. feature level).

\section{Challenges in Temporal Multimodal XAI}
Despite advances, significant challenges remain, including:
\begin{itemize}[parsep=0.5em, itemsep=1em]
    \item \textbf{Architectural Fragmentation:} Models are often specialised to specific domains, hindering comparison and generalisation.
    \item \textbf{Scalability:} Explanation generation can be computationally intensive for long sequences.
    \item \textbf{Evaluation:} Assessing explanation quality is difficult due to the lack of ground truth.
    \item \textbf{Human Factors:} Poorly presented explanations can overwhelm users.
\end{itemize}
Our T-MoXAI framework aims to address these challenges through a unified architecture, efficient computation, hierarchical organisation, and a modular design.

\chapter{Methodology}

\section{Problem Formulation}
We formally define the temporal multimodal learning and explanation problem. Consider a dataset $\mathcal{D} = \{(\mathbf{X}_i, y_i)\}_{i=1}^N$ containing $N$ samples. Each sample consists of a temporal sequence $\mathbf{X}_i$ and an associated outcome $y_i$. Each temporal sequence contains observations at $T_i$ timesteps:
$$ \mathbf{X}_i = \{(\mathbf{x}_{i,t}^{\text{img}}, \mathbf{x}_{i,t}^{\text{tab}}, t)\}_{t=1}^{T_i} $$
where $\mathbf{x}_{i,t}^{\text{img}} \in \mathbb{R}^{H \times W \times C}$ is image data, $\mathbf{x}_{i,t}^{\text{tab}} \in \mathbb{R}^{D}$ is a $D$ dimensional tabular feature vector, and $t$ is the timestamp. Sequences may have variable length ($T_i$) and irregular sampling. Our objectives are to learn a predictive function $f: \mathbf{X} \rightarrow y$ and to generate a hierarchical explanation $E = \{E_{\text{temporal}}, E_{\text{modality}}, E_{\text{feature}}\}$.

\begin{figure*}[t]
    \centering
    \includegraphics[width=0.9\textwidth]{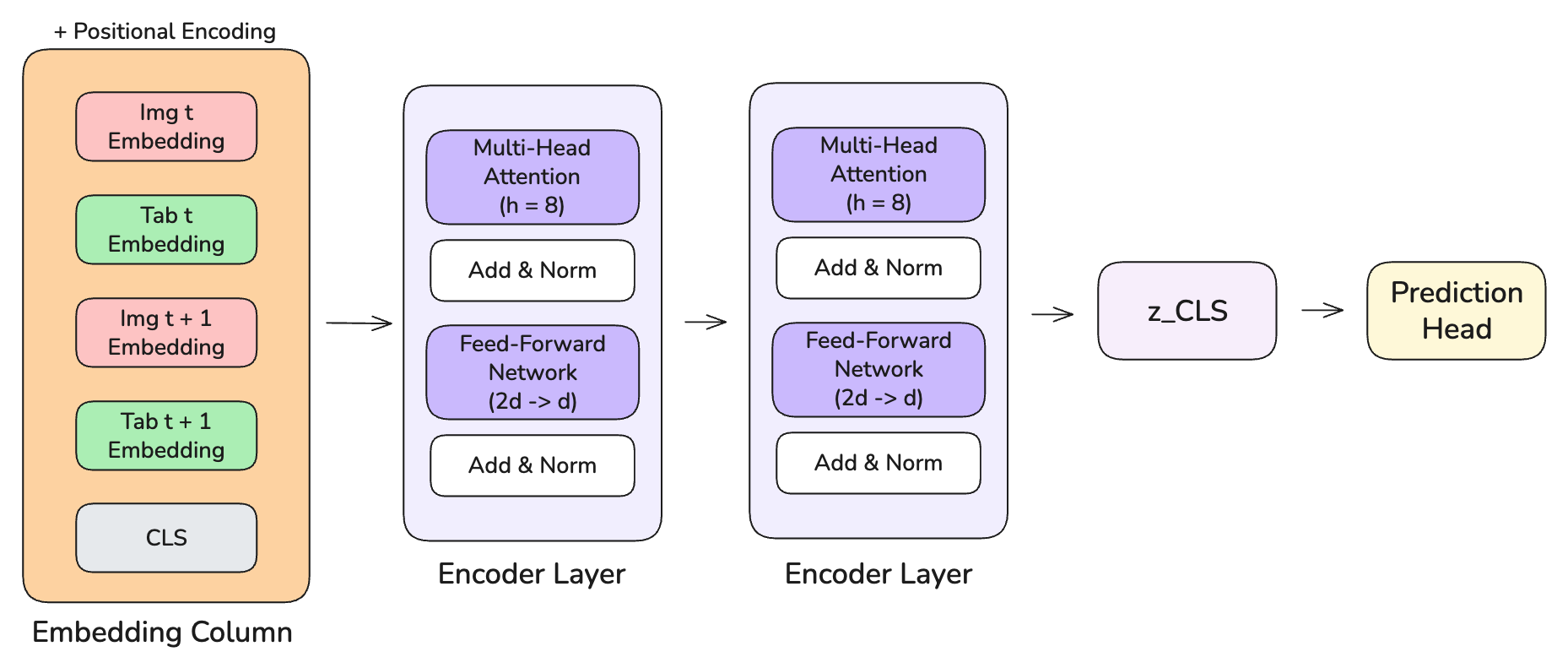}
    \caption{The T-MoXAI architecture. Raw image and tabular data from each timestep are processed by modality specific encoders (ResNet 50 and MLP). The resulting embeddings are interleaved, combined with positional encodings, and fed into a Transformer Encoder. The final prediction is derived from the [CLS] token's output representation.}
    \label{fig:architecture}
\end{figure*}

\section{T-MoXAI Architecture}
The T-MoXAI framework comprises a temporal multimodal prediction model and a hierarchical explanation generator. A diagram of the architecture is shown in Figure~\ref{fig:architecture}.

\subsection{Modality Specific Encoders}
We employ dedicated encoders to transform raw inputs into unified representations.
\textbf{Image Encoder:} We use a ResNet 50 architecture, pretrained on ImageNet, to process visual data:
\begin{equation}
    \mathbf{h}_{i,t}^{\text{img}} = f_{\text{img}}(\mathbf{x}_{i,t}^{\text{img}}; \theta_{\text{img}})
\end{equation}
where $f_{\text{img}}$ is the ResNet 50 backbone and $\mathbf{h}_{i,t}^{\text{img}} \in \mathbb{R}^{512}$ is the encoded representation.
\textbf{Tabular Encoder:} A multi layer perceptron (MLP) with two hidden layers (256 and 512 units, ReLU, dropout p=0.1) processes structured features:
\begin{equation}
    \mathbf{h}_{i,t}^{\text{tab}} = f_{\text{tab}}(\mathbf{x}_{i,t}^{\text{tab}}; \theta_{\text{tab}})
\end{equation}
The output is a 512 dimensional representation matching the image embeddings.

\subsection{Temporal Fusion via Transformers}
We use a transformer to capture temporal patterns and cross modal interactions.
\textbf{Sequence Construction:} We interleave modality embeddings and prepend a learnable `[CLS]` token:
\begin{equation}
    \mathbf{H}_i = [\text{CLS}, \mathbf{h}_{i,1}^{\text{img}}, \mathbf{h}_{i,1}^{\text{tab}}, \dots, \mathbf{h}_{i,T_i}^{\text{img}}, \mathbf{h}_{i,T_i}^{\text{tab}}]
\end{equation}
\textbf{Positional Encoding:} We use learnable positional embeddings that encode both temporal position and modality type to handle irregular sampling.
\textbf{Transformer Architecture:} The sequence is processed by a 4 layer transformer encoder with 8 attention heads per layer.
\begin{equation}
    \mathbf{Z}_i = \text{TransformerEncoder}(\mathbf{H}_i + \mathbf{P})
\end{equation}
\textbf{Task Specific Prediction:} The final prediction is derived from the `[CLS]` token's representation, $\mathbf{z}_i^{\text{CLS}}$, passed through a final linear layer.
\begin{equation}
    \hat{y}_i = f_{\text{pred}}(\mathbf{z}_i^{\text{CLS}})
\end{equation}

\section{Hierarchical Explanation Generation}
T-MoXAI generates explanations at three hierarchical levels.
\subsection{Level 1: Temporal Importance via Shapley Values}
To quantify each timestep's contribution, we adapt Shapley values for the temporal domain. For a sequence with $T$ timesteps, the Shapley value $\phi_t$ for timestep $t$ is:
\begin{equation}
    \phi_t = \sum_{S \subseteq T \setminus \{t\}} \frac{|S|!(T-|S|-1)!}{T!} [v(S \cup \{t\}) - v(S)]
\end{equation}
where $v(S)$ is the model's prediction given only the timesteps in subset $S$. To maintain inputs within the model's expected distribution during this computation, we employ learned mask tokens rather than zero-masking for missing timesteps. For efficiency, we use exact computation for $T \le 5$ and a Monte Carlo approximation otherwise.

\subsection{Level 2: Modality Importance via Attention Analysis}
For each influential timestep, we determine modality importance by analysing the transformer's attention patterns. The importance of modality $m$ at timestep $t$ is the average attention paid to its token by the `[CLS]` token across all layers and heads.

\subsection{Level 3: Feature Level Attribution}
For top influential timesteps, we generate fine grained explanations.
\textbf{Image Explanations via Grad-CAM:} We apply Grad-CAM to identify salient regions in images:
\begin{equation}
    L_{\text{Grad-CAM}}^c = \text{ReLU}\left(\sum_k \alpha_k^c A_k\right)
\end{equation}
where $A^k$ is the $k$-th feature map in the final convolutional layer and $\alpha_k^c$ is its gradient based weight.
\textbf{Tabular Explanations via Integrated Gradients:} For tabular features, we compute Integrated Gradients to obtain reliable attributions:
\begin{equation}
    IG_j(x) = (x_j - x'_j) \int_{\alpha=0}^{1} \frac{\partial f(x' + \alpha(x - x'))}{\partial x_j} d\alpha
\end{equation}
where $x'$ is a baseline input. This method satisfies key axioms like sensitivity and implementation invariance.

\chapter{Experimental Setup}

\section{Datasets}
We evaluate T-MoXAI on two distinct, real world datasets selected to test its domain agnostic capabilities.
\begin{itemize}
    \item \textbf{IVF Outcome Prediction:} A private clinical dataset comprising 1,502 temporal records from IVF treatment cycles, obtained through a collaboration with the Computational Biology Group at the University of Cambridge and the Lister Fertility Clinic. Each patient sequence contains 2 to 8 timesteps corresponding to monitoring visits. Each timestep includes a 2D transvaginal ultrasound image for assessing follicular development and a vector of tabular clinical data (e.g., patient age, BMI, hormone levels, endometrial thickness). The task is a binary classification to predict the outcome of embryo transfer. The dataset exhibits the characteristic class imbalance observed in reproductive medicine, with an overall positive prevalence of 23.1\%. This natural class distribution is preserved, as maintaining realistic outcome probabilities is crucial for clinical decision support and for setting appropriate patient expectations.
    \item \textbf{Wheat Yield Forecasting:} A public agricultural dataset tracking wheat development over a growing season across 4,000 distinct crop plots. Each sequence represents a distinct crop plot monitored at multiple growth stages. Each timestep includes a high resolution RGB image from a drone and a corresponding vector of phenotypic measurements (e.g., plant height, canopy cover). The task is a regression problem to predict the final grain yield. This task is characterised by substantial environmental variability due to weather and soil conditions.
\end{itemize}
For both datasets, we used a 70\%/15\%/15\% split for training, validation, and testing.

\section{Baselines and Implementation Details}
We compare T-MoXAI against a set of baselines to evaluate the contributions of its core components:
\begin{itemize}
    \item \textbf{Last Timestep Only:} A non temporal baseline that uses only the image and tabular data from the final observation in the sequence.
    \item \textbf{Image Only Temporal:} A variant of T-MoXAI that uses only the sequence of images.
    \item \textbf{Tabular Only Temporal:} A variant of T-MoXAI that uses only the sequence of tabular data.
\end{itemize}
All models were trained using the Adam optimiser with a learning rate of $1 \times 10^{-4}$ and a batch size of 16. We used binary cross entropy loss for the IVF classification task and mean squared error loss for the wheat yield regression task. During training, we included an attention entropy regularisation term ($\lambda = 0.05$) to encourage diverse attention patterns and prevent the model from collapsing its focus onto single dominant features. Models were trained until the validation loss did not improve for 10 consecutive epochs.

\spanblock{%
    \begin{tabular}{lcc}
        \toprule
        \textbf{Model} & \textbf{IVF (AUC)} & \textbf{Wheat ($R^2$)} \\
        \midrule
        Last Timestep Only & 0.500 & 0.000 \\
        Image Only Temporal & 0.631 & 0.008 \\
        Tabular Only Temporal & 0.601 & 0.235 \\
        \midrule
        \textbf{T-MoXAI (Ours)} & \textbf{0.660} & \textbf{0.265} \\
        \bottomrule
    \end{tabular}
    \captionof{table}{Predictive performance on the two datasets. T-MoXAI is compared against several baselines (AUC for IVF, $R^2$ for Wheat).}
    \label{tab:performance}
}

\spanblock{%
    \includegraphics[width=\textwidth]{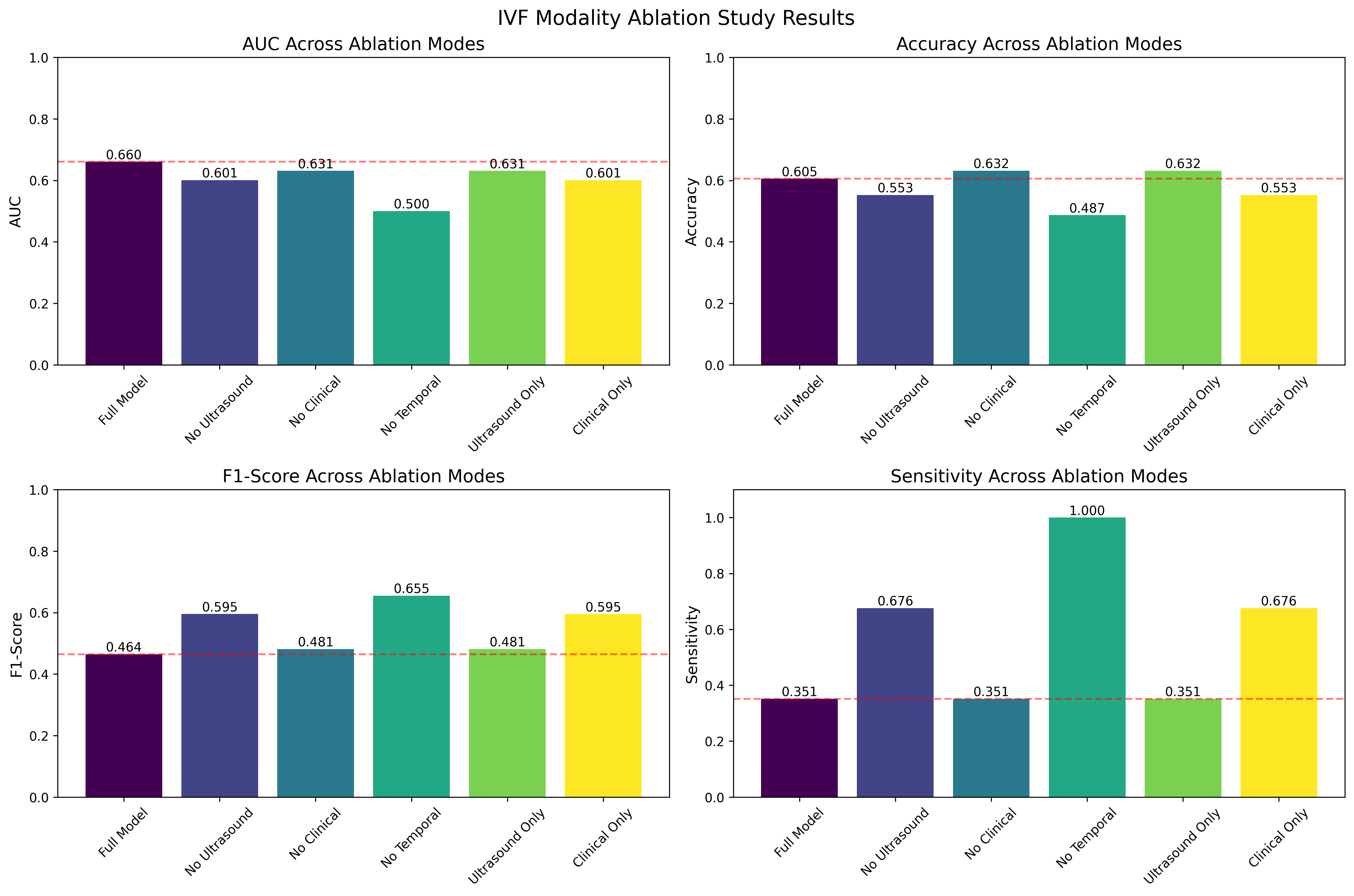}%
    \captionof{figure}{IVF Modality Ablation Study Results. The charts show the impact on AUC, Accuracy, F1 Score, and Sensitivity when key components are removed from the model. The 'No Temporal' case demonstrates the critical importance of temporal modelling.}
    \label{fig:ablation_results}
}

\chapter{Results}

\section{Predictive Performance}
On both tasks, T-MoXAI demonstrates effective performance and outperforms the selected baselines, as shown in Table~\ref{tab:performance}. Our framework achieves an AUC of 0.660 on the imbalanced IVF dataset and an $R^2$ of 0.265 on the noisy wheat yield dataset. These results indicate that the model successfully isolates meaningful physiological and morphological markers despite the noise inherent in clinical ultrasound and variable agricultural field conditions. The performance relative to the baselines suggests the benefits of integrating temporal and multimodal information through the transformer architecture.

\section{Ablation Studies}
To validate our design choices, we conducted several ablation studies. The results for the T-IVF dataset are visualised in Figure~\ref{fig:ablation_results}. As shown in the AUC plot, the full model achieves the highest performance (0.660). Removing either the ultrasound modality (AUC 0.601) or the clinical modality (AUC 0.631) degrades performance, indicating that both provide complementary information. Most critically, removing the temporal modelling component entirely causes performance to drop to a random chance level (AUC 0.500), underscoring that temporal awareness is essential for this task. A similar pattern was observed on the Wheat yield task, where removing temporal modelling also resulted in model failure ($R^2 < 0.00$), and both modalities were found to contribute to the final performance.

\section{Qualitative Explanation \\Analysis}
Qualitative analysis of individual predictions reveals the nuances of the model's decision process. For instance, in a true positive IVF case (Figure \ref{fig:explanation_example}), temporal Shapley values show that a mid-cycle visit was most influential, while information from earlier visits was actively discounted. The corresponding Grad-CAM explanation is particularly revealing; it highlights regions outside the primary anatomical structures of interest. This suggests the model may be using contextual information or image artifacts, demonstrating how the explanation framework can uncover unexpected model behaviours that warrant further investigation.

Similarly, for agricultural yield forecasting, T-MoXAI demonstrates sophisticated temporal and domain-specific reasoning. In a high-yield prediction scenario (Figure \ref{fig:fip_explanation_example}), the model placed extreme importance on early establishment (day 0) and mid-season vigour (day 28), reflecting the biological reality that successful early canopy establishment and peak vegetative growth prior to anthesis are critical determinants of final grain yield. The corresponding Grad-CAM saliency maps appropriately focused on crop material rather than soil or shadows.

\spanblock{%
    \includegraphics[width=0.9\textwidth]{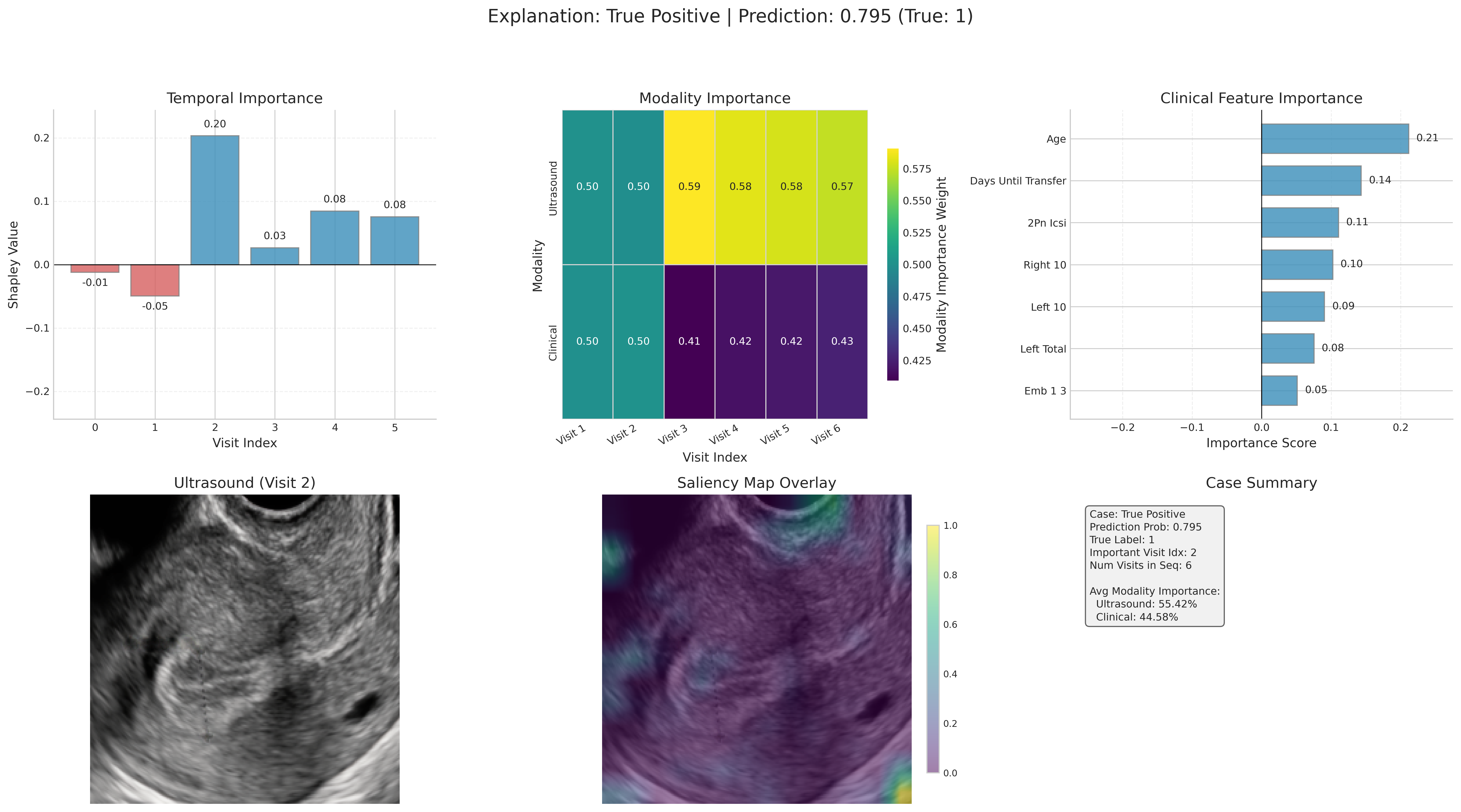}
    \captionof{figure}{Example of a three level explanation for a single IVF prediction. (a) Temporal importance (Shapley values). (b) Modality importance (attention weights). (c) Feature attribution (Grad-CAM).}
    \label{fig:explanation_example}
}

\section{Faithfulness Evaluation}
We evaluated the faithfulness of our temporal explanations using a temporal adaptation of the ROAR (Remove and Retrain) method. We compared the performance degradation when iteratively removing timesteps in descending order of their Shapley value importance versus removing them in a random order. The model's performance declined significantly faster when removing the most important timesteps first. Specifically, for the T-IVF dataset, masking the top 3 most important visits caused the AUC to drop by 0.042, compared to a drop of only 0.027 for random masking (yielding a faithfulness ratio of 1.58). For the FIP 1.0 dataset, masking the top 3 timepoints increased the Mean Squared Error (MSE) by 0.112, versus an increase of just 0.018 for random masking (achieving a capped faithfulness ratio of 3.0). These quantitative results provide strong evidence that the explanations generated by T-MoXAI faithfully reflect the parts of the input sequence that are most influential to the model's decision-making process.

\end{multicols}
\clearpage
\begin{multicols}{2}

\spanblock{%
    \includegraphics[width=0.9\textwidth]{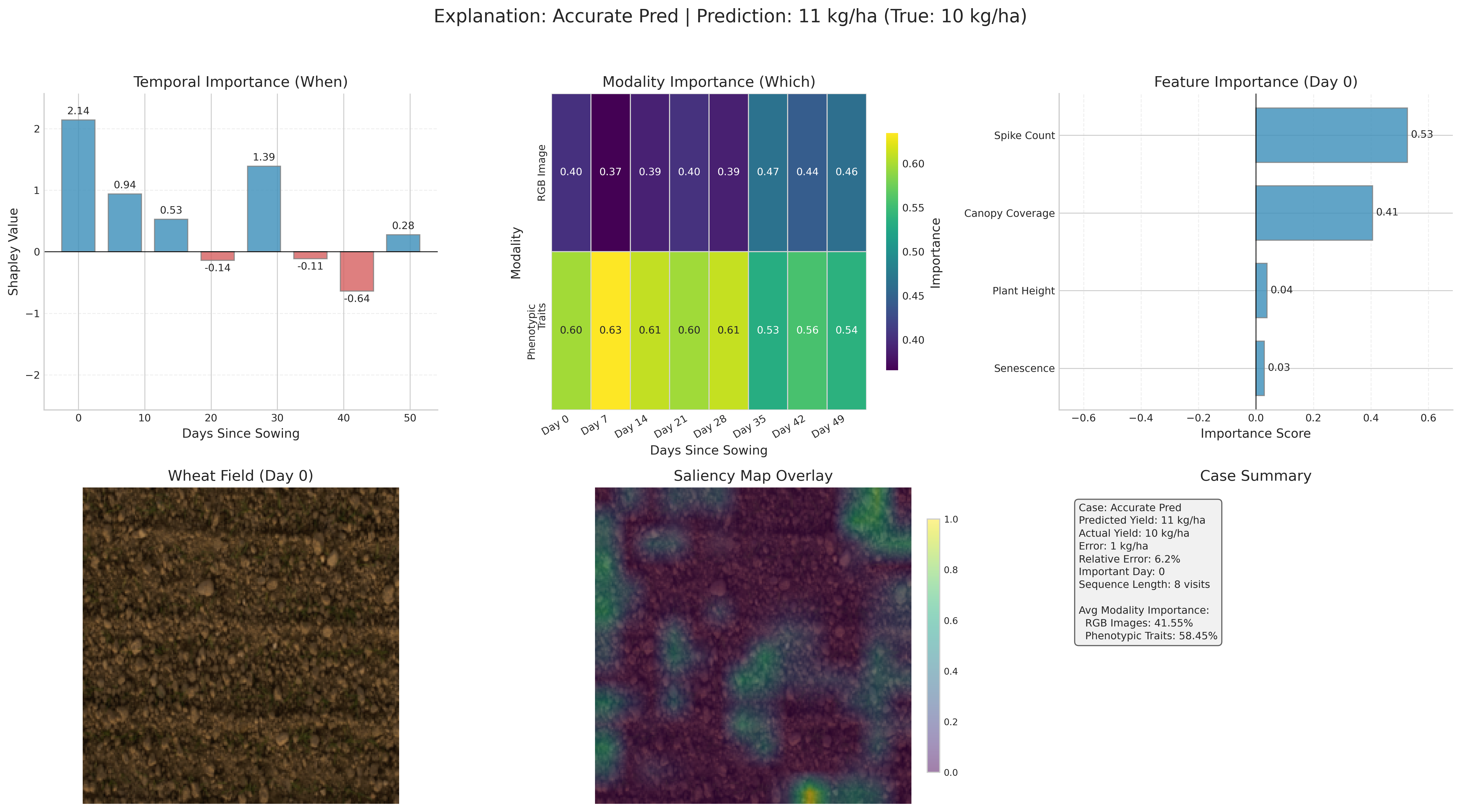}
    \captionof{figure}{Example of a hierarchical explanation for a high-yield wheat prediction from the FIP 1.0 dataset. The framework correctly identifies critical early-season and mid-season growth stages as highly influential, shifting attention between visual and phenotypic data as the crop matures.}
    \label{fig:fip_explanation_example}
}

\nobreak

\chapter{Discussion}

\section{Limitations}
Our results suggest a need for methods that can improve model sensitivity while maintaining well calibrated probability estimates, which are important for clinical decision making. The maximum sequence length of 16 reflects computational constraints but may not capture all relevant long term patterns, as some agricultural studies span entire seasons and fertility treatments can extend over multiple cycles. Developing efficient methods for very long sequences remains an open challenge. While our computational faithfulness metrics provide initial validation, they cannot definitively assess the practical utility of the explanations in human decision making. The lack of ground truth for explanations and the cost of expert evaluation remain fundamental challenges in XAI. Future work could include structured user studies with domain experts to assess practical utility.

\section{Broader Implications}
The current fragmentation in temporal multimodal AI, where different studies often use varied architectures and datasets, can hinder direct comparison and progress. By providing an open source, domain agnostic framework, we aim to offer a baseline for standardised evaluation. Our results suggest that integrating explainability into a system's design does not necessarily compromise performance. The marginal computational overhead for generating explanations (sub second) is small relative to the potential gains in trust and insight. The observed similarities in temporal patterns across domains, with both IVF and agriculture showing critical early phases, may suggest potential for transfer learning.

\section{Future Research Directions}
Future work could explore interactive systems that allow users to drill down from high level temporal importance to specific features. Integrating uncertainty quantification, perhaps through Bayesian approaches or ensembles, could produce richer explanations. Developing methods for temporal counterfactual reasoning could provide actionable insights for optimising monitoring protocols. Furthermore, the success of large language models suggests potential for foundation models trained on vast temporal multimodal datasets, which could provide strong representations for downstream tasks.

\section{Practical Deployment Considerations}
Deploying T-MoXAI in clinical settings would require careful attention to workflow integration. Explanations should be presented in familiar terminology, and systems must handle missing data gracefully. As medical AI faces increasing scrutiny, explainability may become a regulatory expectation; our hierarchical explanations could provide auditable reasoning trails. Beyond decision support, T-MoXAI's explanations could also serve educational purposes for junior clinicians or in agricultural extension services.

\chapter{Conclusion}
This work presented T-MoXAI, a hierarchical explainability framework designed to address the need for interpretable AI systems that process temporal multimodal data. Through evaluation on real world healthcare and agricultural datasets, we show that it is possible for complex AI systems to provide explanations without sacrificing predictive utility. Our contributions include a unified, domain agnostic architecture; hierarchical three level explanations that align with expert reasoning; cross domain validation achieving an AUC of 0.660 for IVF prediction and an $R^2$ of 0.265 for wheat yield forecasting; and an efficient implementation suitable for interactive use.

Temporal modelling was found to be essential, with its removal causing a collapse in model performance in both domains. The model's integration of modalities dynamically adapted to the specific task; IVF predictions balanced ultrasound and clinical data, whereas agricultural yield forecasts relied predominantly on quantitative phenotypic traits. Hierarchical explanations successfully isolated relevant clinical structures and critical phenological transitions, while computational faithfulness evaluations confirmed that these highlighted timesteps genuinely drive the model's predictive output.

The implications of this work extend beyond its technical contributions. By providing interpretable insights, T-MoXAI aims to help bridge the gap between AI capabilities and human trust. Our open source implementation also offers a standardised baseline to facilitate future research. Significant challenges remain, including data limitations and scaling to longer sequences. Future work should explore larger scale data collection, uncertainty aware explanations, interactive interfaces, and foundation models for temporal multimodal representation learning. As AI systems tackle increasingly complex decisions, frameworks like T-MoXAI become more important for ensuring these tools remain understandable and aligned with human reasoning.

\end{multicols}
\balance 

\clearpage

\bibliographystyle{unsrt}
\bibliography{references} 

\end{document}